\pdfoutput=1

\documentclass[11pt]{article}

\usepackage[]{ACL2023}

\usepackage{times}
\usepackage{latexsym}

\usepackage[T1]{fontenc}

\usepackage[utf8]{inputenc}
\usepackage[nameinlink]{cleveref}
\usepackage[]{color-edits}
\addauthor{ZT}{olive}

\usepackage[]{todonotes} %

\usepackage{microtype}
\usepackage{csquotes}
\usepackage{inconsolata}
\usepackage{booktabs}
\usepackage{graphicx}
\usepackage{tabularx}
\usepackage{multirow}
\usepackage{enumitem}

\usepackage{soul}

\newcommand*\samethanks[1][\value{footnote}]{\footnotemark[#1]}

\title{Understanding ``Democratization'' in NLP and ML Research}

\author{Arjun Subramonian\thanks{~~Equal contribution.}~~\textsuperscript{1} \,
 Vagrant Gautam\samethanks~~\textsuperscript{2} \\ \textbf{Dietrich Klakow\textsuperscript{2}} \,
 \textbf{Zeerak Talat\textsuperscript{3}} \vspace{3px} \\ \vspace{3px}
 \textsuperscript{1}University of California, Los Angeles, USA~~~ 
\textsuperscript{2}Saarland University, Germany \\
 \textsuperscript{3}Mohamed Bin Zayed University of Artificial Intelligence, UAE~~~\\
 \small{\tt  arjunsub@cs.ucla.edu} \\
\small{\tt  \{vgautam, dietrich.klakow\}@lsv.uni-saarland.de} \\
\small{\tt  z@zeerak.org} \\
}

\begin{document}
\maketitle

\begin{abstract}
  Recent improvements in natural language processing (NLP) and machine learning (ML) and increased mainstream adoption have led to researchers frequently discussing the ``democratization'' of artificial intelligence.
  In this paper, we seek to clarify how democratization is understood in NLP and ML publications, through large-scale mixed-methods analyses of papers using the keyword ``democra*'' published in NLP and adjacent venues. %
  We find that democratization is most frequently used to convey (ease of) access to or use of technologies, without meaningfully engaging with theories of democratization, while research using other invocations of ``democra*'' tends to be grounded in theories of deliberation and debate.
  Based on our findings, we call for researchers to enrich their use of the term democratization with appropriate theory, towards democratic technologies beyond superficial access.\footnote{Our code is available at:
  \url{https://github.com/ArjunSubramonian/democratization-nlp}.}

\end{abstract}

\section{Introduction}
\begin{figure}[t!]
    \centering
    \includegraphics[width=\columnwidth]{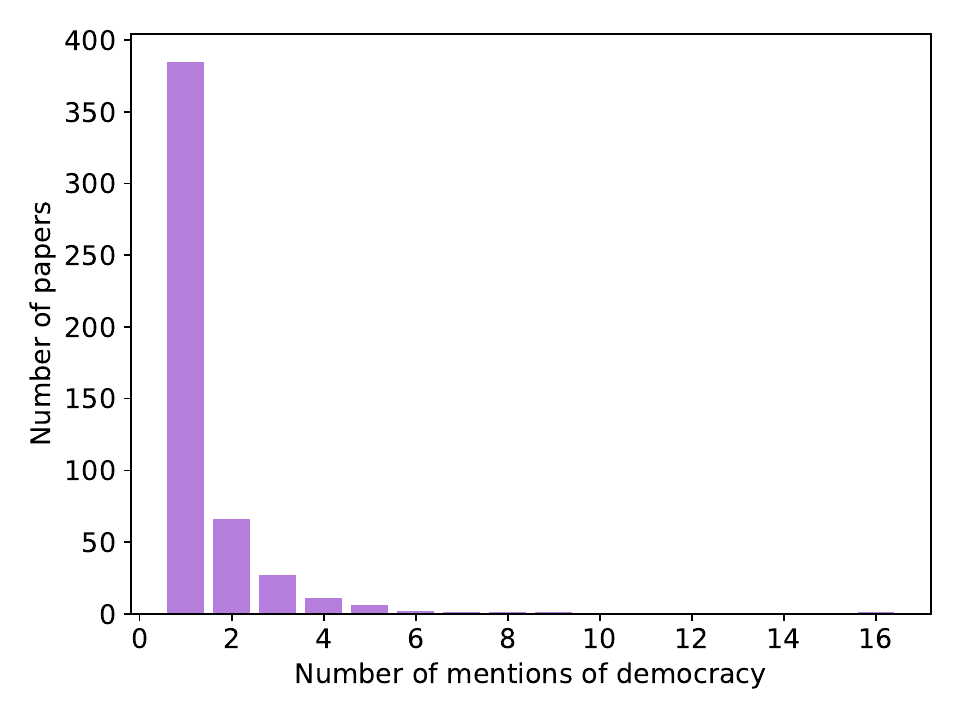}
    \caption{Frequency of mentions of democracy ($>$0) per paper in all work published in the ACL Anthology, ICLR, ICML, or NeurIPS before November 24, 2023.
    76.1\% of papers only mention democracy once.}
    \label{fig:mentions-per-paper}
\end{figure}

As the influence of language technologies has grown, %
it has become increasingly popular to discuss ``democratization'' in natural language processing (NLP) and machine learning (ML) research~\citep{Seger2023DemocratisingAM};
for instance, OpenAI has invested in a ``democratic process for deciding what rules AI systems should follow''~\cite{Zaremba_Dhar_Ahmad_Eloundou_Santurkar_Agarwal_Leung_2023}, Anthropic has explored how ``democratic processes can influence artificial intelligence (AI) development''~\cite{AnthropicAI}, and HuggingFace has stated their mission to be to ``democratize good machine learning''~\cite{HuggingFace}.
Indeed, a large number of NLP and ML papers mention terms related to democracy (see \Cref{fig:mentions-per-paper}), thereby raising the question: What do we understand by ``democracy'' and ``democratization'' when we invoke them in research?

So far, the treatment of democracy in NLP and ML literature, and particularly the term ``democratization,'' has not been subject to careful investigation.
Our paper fills this gap by analyzing uses of ``democratization'' %
in NLP and ML papers, and %
their connections to democracy.
We examine conceptualizations of these terms through a large-scale mixed-methods analysis of every use of ``democra*'' in papers published in the Anthology of the Association of Computational Linguistics (ACL Anthology), the International Conference on Learning Representations (ICLR), the International Conference on Machine Learning (ICML), and Neural Information Processing Systems (NeurIPS). %

We find that on one hand, the use of ``democratization'' tends to indicate a broadening of access to research artifacts, particularly without domain expertise, while NLP and ML literature discussing democracy in other contexts is often rooted in theories of deliberation and debate. We also find that while authors associate democratization with positive values related to access and reducing costs, the term itself is rarely defined or operationalized.
Prior work has argued that the ``democratization of AI revolves primarily around the notion of access'' \citep{Burkhardt2019, Sudmann2019,Sudmann2019-2, Luchs2023}; our work provides systematic evidence for this claim, and is grounded in a comparison to other democracy-related terms.

Next, we examine papers that mention democracy for their depth of engagement with the topic, by exploring their text and citations.
We find that a majority of papers only invoke democracy once, do so outside of methods and results sections, and engage minimally with extra-disciplinary work.

We conclude that ``democratization'' constitutes a misnomer for ``access,'' and therefore encourage future work to either enrich their research by drawing on over 3000 years of scholarship on democracy and democratization, or  use ``access'' instead.
Lacking clear, consistent and responsible use of the term ``democratization,'' NLP and ML risk misrepresenting progress in %
capturing democratic values, the distribution of power, and public control of AI.
Clearer conceptualizations of ``democratization'' can thus strengthen progress towards truly democratic technologies beyond just superficial access.

\section{Related Work}

\subsection{Democratization beyond AI}
\label{sec:democratization-beyond-ai}

The use of ``democratization'' extends far beyond AI.
In conservation biology, for instance, it is discussed in the context of citizen and community science;
public participation in processes such as data collection is seen as democratizing knowledge production~\citep{Bela2016LearningAT}, and reducing gaps between academia and wider society~\cite{Sauermann2020CitizenSA}.
However, constraining community science to participation has also been criticized as ``participation washing''~\cite{sloane2022participation}, as it often disregards local knowledge, prevents the public from formulating scientific questions, and fails to change the norms of institutions~\cite{Kimura2016CitizenSP}.
In contrast, political scientists examine the democratization of policy research through ``collaborative citizen-expert inquiry''~\cite{Fischer1993CitizenPA} which has been considered essential to democratically tackling social issues~\cite{Weinberg_2022}.
Internet scholars investigate the democratizing effects of online information and social media, i.e., how they have helped to spread pro-democratic ideas, discussions, and protests globally~\cite{Hill1999IsTI,Weinstein2012TheDO}.
Beyond research, there have been calls towards protecting the integrity of democracy through the democratization of media and ``free access to pluralistic information and opinion''~\cite{news-democratization}.
In relation to emerging democracies, the democratization of media is often linked to the diversification of news sources~\cite{Barnett1999TheLO, Tettey2001TheMA, Porto2012MediaPA}.

\subsection{Conceptions of democratization in AI}
\label{sec:democratization-in-ai}

Research in AI has presented access-centric conceptions of democratization,
e.g., to identify criteria for democratizing the use of AI, such as affordability, accessibility, and fairness \citep{Ahmed2020AFF}.
Similarly, \citet{Ahmed2020TheDO} conceptualize democratization as equity in access to compute between tech companies and non-elite universities.
However, this line of research has not examined the possible connections between democratization and democracy.
Prior work has also challenged the conceptualization of democratization in AI. \citet{Seger2023DemocratisingAM} argue that disparate uses of the term ``democratization'' have caused a lack of recognition of shared ``goals, methodologies, risks, and benefits.''
Drawing from news articles and talks, they identify four notions of democratization: use, development, benefits, and governance.
Similarly, in a study of 35 articles on the use of ``democratization'' and its connection to democracy within the scope of medical AI, \citet{Rubeis2022DemocratizingAI}  uncover diverse conceptualizations, from increasing data access to AI governance.

Another line of work, focusing on AI governance and increased public control of AI development and deployment, argues that public participation is critical for democratizing AI, e.g., \citet{Gilman2023Democratizing} calls for institutions to budget for participation at all stages of AI development.
Participation has also been operationalized by aligning models to a ``constitution'' based on the values of human representatives~\cite{Siddarth2023AIDem}; 
by connecting open-source and democratic communities, and widening geographic diversity in public input processes~\cite{CIP2024Roadmap}; and by leveraging ``democratic'' frameworks to gather AI uses, harms, and benefits from the public to guide the evaluation and regulation of AI~\cite{mun2024particip}.
However, these approaches offer minimal opportunities for publics to %
contest the logics and power structures of the AI industry~\cite{Luchs2023}.

In contrast to these bodies of work, we perform a large-scale mixed-methods analysis of papers published at NLP and ML venues. %
Similarly to \citet{Seger2023DemocratisingAM} and \citet{Rubeis2022DemocratizingAI}, we find distinct conceptualizations of democratization that obviate its benefits and risks, often due to a lack of theoretical engagement.
Ultimately, our analysis shows that the dominant conception of democratization is access, and that a shared understanding of democratization and democracy, which is essential for democratic frameworks, remains absent within the NLP and ML community at large.

\begin{table*}[ht!]
    \centering
    \begin{tabularx}{\linewidth}{Xcc}
    \toprule
    \multicolumn{1}{c}{\textbf{Excerpt}} & \multicolumn{1}{c}{\textbf{Themes}} & \multicolumn{1}{c}{\textbf{Concepts}} \\
    \midrule
        \textit{``The right to access judicial information is a fundamental component of Canadian democracy and its judicial process.''} & \multirow{2}{6em}{\centering necessary / beneficial} & \multirow{2}{6em}{\centering access, information} \\
        \midrule
        \textit{``An abundance of incorrect information can plant wrong beliefs in individual citizens and lead to a misinformed public, undermining the democratic process.''} & \multirow{3}{*}{danger} & \multirow{3}{6em}{\centering citizenship, misinformation} \\
        \midrule
        \textit{``This is a totally democratic method where each vote counts the same.''} & \multirow[c]{2}{*}{math} & \multirow[c]{2}{*}{equal contribution} \\
        \midrule
        \textit{``This helps to improve data literacy, democratizing accessibility to otherwise opaque public database systems.''} & \multirow[c]{2}{*}{democratization} & \multirow[c]{2}{*}{access, data} \\
         \bottomrule
    \end{tabularx}
    \caption{Example excerpts each of the four themes, along with the associated concepts we annotate.}
    \label{tab:themes-concepts-values}
\end{table*}

\section{Data}

Using the Semantic Scholar API~\cite{kinney2023semantic}, we collect all papers published before November 24, 2023 in the ACL Anthology, ICML, ICLR, and NeurIPS, that mention terms related to ``democracy.''
We choose these venues, as they are top-tier NLP and ML conferences that influence practices in the field.
We obtain 1,537 papers, which we filter for relevance, obtaining a final dataset of 506 papers and 916 excerpts for analysis.

\paragraph{Obtaining Excerpts}
We first collect all metadata and text from open-access PDFs using the Semantic Scholar API.
We split the text of each paper using the punkt NLTK sentence tokenizer~\cite{bird-loper-2004-nltk}, and extract all sentences that contain the substring ``democra'' (excluding ``democrats''), resulting in 4,203 excerpts across 1,709 papers.
We do not include related terms (e.g., participatory governance, constitution, etc.) so that we do not inadvertently select irrelevant papers, and to keep our discussion firmly grounded in a comparison between democratization and democracy.

\paragraph{Filtering Irrelevant Excerpts} %

 In order to identify excerpts that reveal how authors conceptualize ``democratization'' and ``democracy,'' we remove unrelated uses of ``democra,'' such as those in named entities (e.g., ``Center for Media and Democracy''), motivating examples (e.g., for textual entailment), modeling examples (e.g., LDA topics), examples from datasets (e.g., tweets), mentions in non-English languages, and references.
We perform this filtering using a two-stage approach: automatic filtering and manual annotation for relevance.

We curate a list of terms (see \Cref{sec:methodological-details}) for automatically filtering excerpts:
We exclude named entities (e.g., ``the Syrian Democratic Forces'') and terms that exclusively appear as examples of data (e.g., tweets containing ``\#democracy'').
One author verified all automatically filtered excerpts.

After filtering, we manually annotate the remaining 2,273 excerpts, searching for instances where the authors deliberately use words containing ``democra'' as part of their argument or evidence, examining the full PDF in ambiguous cases.
After concluding the two-stage filtering process, we obtain 916 excerpts from 506 papers for analysis.

\section{Conceptualizations of Democracy}
\label{sec:conceptualizations}

To understand how democracy and democratization are conceptualized by authors in NLP and ML papers, we inductively analyze our data for overarching themes, values, and concepts. %
We find that conceptualizations of democratization are distinct from democracy, and instead are closely related to access and financial costs.

\subsection{Methodology}
Two authors annotate the first 300 excerpts independently for themes, concepts, and values in an open-ended manner (see \Cref{tab:themes-concepts-values} for example excerpts and annotations).
We then resolve inconsistencies and consolidate themes, concepts, and values, before annotating the remaining excerpts independently.
Finally, we group the themes, concepts, and values, respectively,
into sets per paper.

\paragraph{Themes}
We %
qualitatively code %
the excerpts to identify salient, overarching themes that characterize how they discuss democracy; this is a common inductive methodology from the social sciences described by \citet{Saldana2021-ki}.
Four major categories emerge after a first pass over all the excerpts:
\begin{itemize}[leftmargin=0.75em]
    \itemsep0em 
    \item \textit{Necessary/Beneficial}: things that are necessary for or beneficial to democracy (e.g., discourse, majority, voting)
    \item \textit{Danger}: dangers to democracy (e.g., misinformation)
    \item \textit{Democratization}: use of the words ``democratize'' or ``democratization'' (e.g., of ML)
    \item \textit{Math}: mathematical or ML ways to operationalize democracy (e.g., democratic matrices, mathematical models of democracy)
\end{itemize}

Two authors then systematically annotate every excerpt with an explicit and, if applicable, an implicit theme. 
An explicit theme is assigned to excerpts that explicitly state, e.g., that something is necessary for or a danger to democracy; otherwise, it is classified as \textit{other}. 
In contrast, the implicit theme requires annotators to make inferences about how researchers think about democracy.

\noindent For example, the excerpt: ``The most democratic option is to give each tagger one vote (Majority),'' is assigned an explicit theme of \textit{math}, as it discusses  operationalizing NLP taggers in a ``democratic'' way. 
We also infer that the authors believe majority voting is necessary for democracy, hence \textit{necessary/beneficial} is assigned as an implicit theme.

We do not differentiate between papers about the effect of democracy on technology (e.g., \textit{danger}) and democratic principles in technology (e.g., \textit{democratization}), as all papers that invoke democracy-related terms can engage with democratic theories, and both themes relate to participation. Not distinguishing between them and instead inductively looking for what patterns emerge allows us to identify how differently democratization and democracy may be conceptualized. %

\paragraph{Values and Concepts}
The same two authors also label each excerpt for values (e.g., ``consensus'' and ``equality'') and more broadly concepts (e.g., ``misinformation'' and ``elections'') associated with democracy to explore conceptualizations of democracy more granularly. 
We focus on values (a subset of concepts) in our main analysis;
see \Cref{sec:additional-results} for further discussion of values and concepts.

\begin{figure}[t!]
    \centering
    \includegraphics[width=\linewidth]{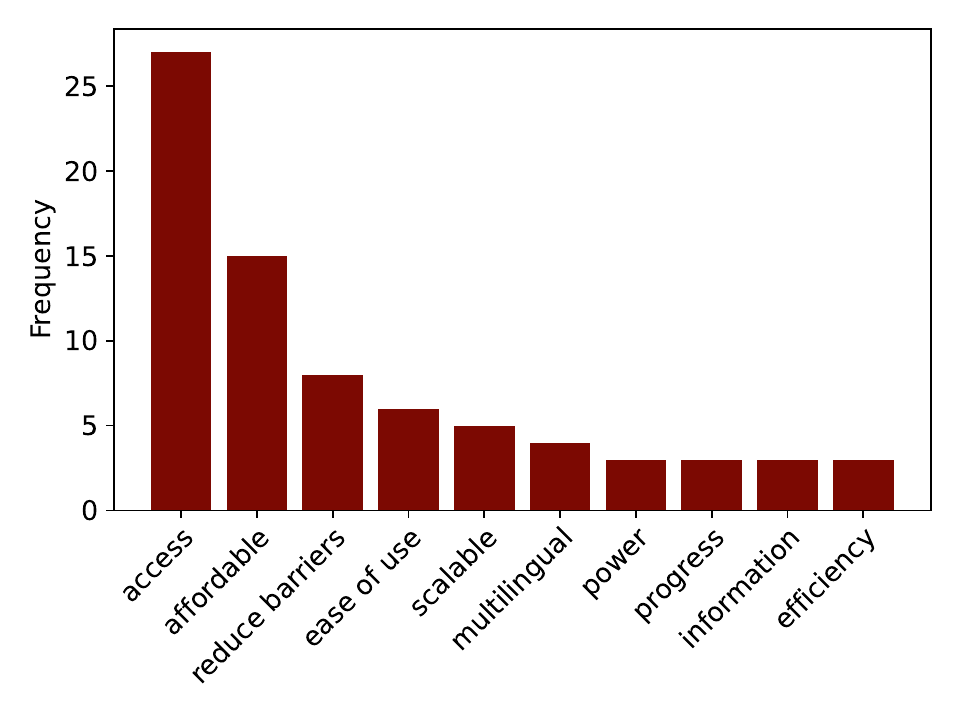}
    \includegraphics[width=\linewidth]{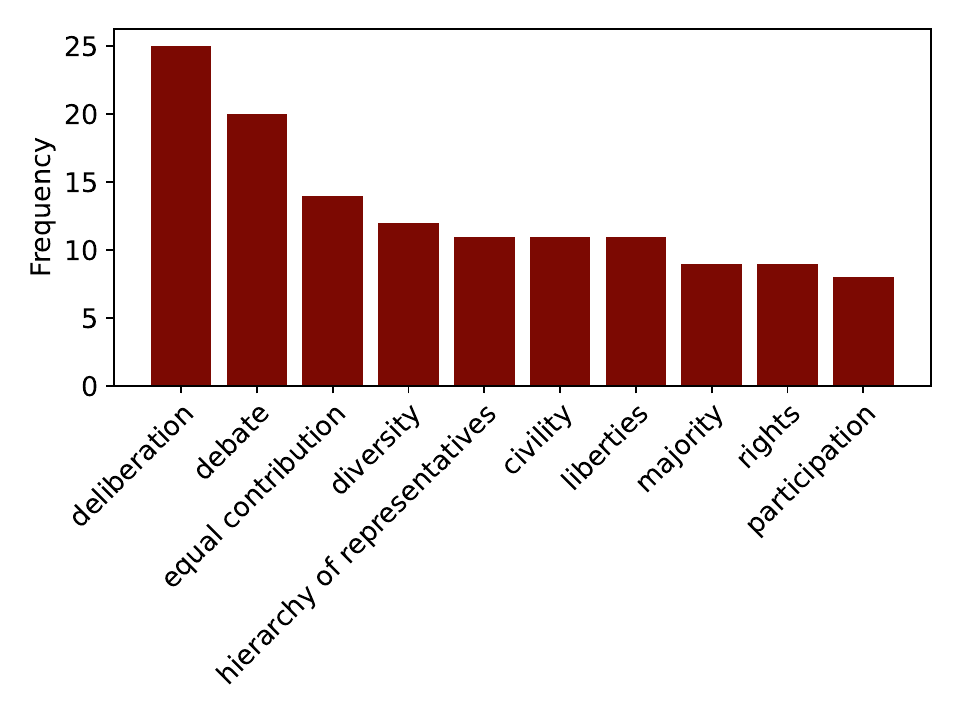}
    \caption{Frequency of values, split by \textit{democratization} papers and all other papers. Associations with \textit{democratization} (top) are different from associations with all other mentions of democracy (bottom).}
    \label{fig:concepts-values}
\end{figure}

\subsection{Results}
Of the four themes, we find that \textit{democratization} is by far the most frequent with 213 papers, followed by 67 for \textit{necessary/beneficial}, 58 for \textit{danger}, and 35 for \textit{math}.
In total, we identify 110 concepts (including 77 values) associated with democracy, with each paper containing an average of 1.16 themes and 1.036 concepts.
For themes, annotation is highly consistent, with a Cohen's kappa of 0.973 for explicit themes and 0.887 for implicit themes using the Jaccard distance metric \citep{Cohen1960ACO}. Given the minimal disagreement between authors, we henceforth do not distinguish between explicit and implicit themes. For concepts, annotators have a Cohen's kappa of 0.349. Although this score only indicates fair agreement in the binary classification setting \citep{McHugh2012InterraterRT}, with 110 possible concepts, there is a much lower random chance of agreement, and thus 0.349 reflects moderate to high agreement in this context.

\paragraph{Values Associated with Democracy in NLP and ML}

\Cref{fig:concepts-values} shows the values associated with ``democratization'' compared to all other mentions of democracy.
We find that some values contradict each other.
For instance, work has conceptualized ``random selection,'' ``consensus,'' and ``majority (voting)'' as democratic, however these are all mutually exclusive of one another.
Yet, researchers conceptualize NLP and ML systems operating in these three manners as ``democratic,'' showing the need to explicitly consider how different systems require different conceptualizations of democracy.

We find that \textit{non-democratization} papers identify values and concepts that readily connect to widespread
theoretical notions of democracy, e.g., decision-making, deliberation, debate, and diversity.
In contrast, \textit{democratization} papers are overwhelmingly associated with increasing access, ease of use, and reducing costs and barriers.
That is, democratization papers share values with radical egalitarian theories of democracy (see \Cref{sec:background}), but do not distinguish or make apparent the relationship between access and equal access to democratic processes.
Thus, in contrast to other fields (see \Cref{sec:democratization-beyond-ai}), NLP and ML researchers who use these words seem to conceive of \textit{democratization} quite differently from democracy, associating them with different and sometimes conflicting values, and agreeing primarily that both are aspirational.

\section{Democratization in NLP and ML}
\label{sec:democratizaton-nlp-ml}

Given that \textit{democratization} in NLP and ML is markedly different from the other themes, we examine the politics of democratization in papers with this theme.
Specifically, we consider \textit{what} is being democratized, \textit{how}, and \textit{to what end}?

\subsection{Methods}

To examine the politics of \textit{democratization}, one author
annotates all excerpts with an explicit theme of \textit{democratization} for \textbf{targets of democratization}, i.e., what the object of democratization is; \textbf{causes of democratization}, i.e., how is an object being democratized, or what engenders its democratization; and the \textbf{goals of democratization}, i.e., why or to what ends an object is being democratized. See example excerpts and annotations in \Cref{tab:dem-causes-targets-goals-examples}.

\begin{table*}[t]
    \centering
    \begin{tabularx}{\linewidth}{Xccc}
        \toprule
        \textbf{Excerpt} & \textbf{Cause} & \textbf{Target} & \textbf{Goal} \\
        \midrule
        \textit{``We aim at an ambitious goal of democratizing the cost of pretraining.''} & & \multirow{2}{*}{cost} & \\
        \midrule
        \textit{``We narrow our purview to open source and accessible data collections, motivated by the goal of democratizing accessibility to research.''} & \multirow{3}{5em}{\centering data} & \multirow{3}{5em}{\centering access, research} & \multirow{3}{*}{access} \\
        \midrule
        \textit{``With everyone being able to create data for their model training, we can pave the way for the democratization of AI.''} &  & \multirow{3}{*}{AI} & \multirow{3}{7em}{\centering access, use without expertise} \\
        \bottomrule
    \end{tabularx}
    \caption{Top causes, targets and goals of democratization in the 213 papers that mention it.}
    \label{tab:dem-causes-targets-goals-examples}
\end{table*}

\subsection{Results}
\label{sec:democratization-results}

\begin{table}[t]
    \centering
    \begin{tabularx}{\linewidth}{cX}
        \toprule
        \textbf{Causes} & \textit{None specified} (59\%), compute reduction, data, cost reduction, social media, time reduction, open source, internet, access, tools, research, model hubs, libraries \\
        \textbf{Targets} & Research, access, NLP, AI, ML, content creation, DL, language models, MT, internet, information, RL, data \\
        \textbf{Goals} & \textit{None specified} (75\%), use without expertise, access, increased language use, social good, reduce barriers, multilingual, sociological phenomena, quality issues, broader audience, fake news, commodification \\
        \bottomrule
    \end{tabularx}
    \caption{Top causes, targets and goals of democratization in the 213 papers that mention it.}
    \label{tab:dem-causes-targets-goals}
\end{table}

We find that 59\% of the papers
do not state causes of democratization and 75\% do not state the goals.
A subset of authors describe democratization as a separate autonomous process that is, at best, minimally affected by their contributions.

Other authors posit that their research democratizes a technology, but do not elaborate on how that occurs, e.g., in terms of digital infrastructure, governance structures, participatory methods, etc.
When stated, popular causes (see \Cref{tab:dem-causes-targets-goals}) for democratization are reductions in required compute, time, and cost.
Targets for democratization are more nebulous; for instance, authors indicate NLP, AI, or research and access are the target for democratization, however what it means for any of these to be democratized is unclear at such a level of abstraction.
In contrast, the primary goals of democratization are increasing access and use, particularly without requiring expertise. 
However, without consideration of the causes and the targets of democratization, such goals appear inherently elusive.

We validate our excerpt-based results by sampling papers for close readings of the entire articles.
We identify set of papers by using the Huggingface~\cite{wolf2020huggingfaces} \texttt{all-mpnet-base-v2} sentence transformer~\cite{reimers-2019-sentence-bert} to embed all excerpts related to \textit{democratization}.
Then, we apply spectral clustering to the embeddings (\Cref{fig:clustering} in \Cref{sec:methodological-details}) and select 3 clusters using the spectral gap heuristic.
We select 5 papers from the cluster centers and 5 from the boundaries from each cluster, for a total of 30 papers.
Our close reading of our
sampled papers confirm our excerpt analysis: none of the selected papers consider \textit{what} is being democratized, or plan for \textit{how} to democratize. 
Indeed, very few even comment on democratization outside of the excerpts.

\section{Engagement with Democratic Theories}
\label{sec:depth-and-breadth-of-engagement}

Given such a lack of consideration within the \textit{democratization} theme, we examine how NLP and ML papers engage with literature on democracy to understand its influence on the conceptualizations in \Cref{sec:conceptualizations}.
We argue that discussing democracy or democratization without connecting to established theories reflects subpar interdisciplinarity and citational praxis, and risks misrepresenting how grounded AI is in democratic values.

\subsection{Methods}
We measure the depth of engagement with democracy by counting where and how often ``democra*'' terms are mentioned in papers.
We extract section names using the Semantic Scholar API and normalize them across papers, e.g., mapping ``Related Works'' to ``Related Work.''
For a complementary view of engagement that is not limited to words containing the substring ``democra,'' we also study the references these papers cite: the fields they belong to,
the proportion of extra-disciplinary citations,
and citational \textit{intent}, i.e., whether the citation is used to provide background, inform the methodology of the paper, or is related to the results. 
This analysis allows us to evaluate engagement with theories of democracy.
We obtain field, venue and intent metadata using the Semantic Scholar API; we classify references as \textit{intra-disciplinary} if they are from Computer Science, Mathematics, or Linguistics, and as \textit{extra-disciplinary} otherwise.
Finally, we confirm the results of our computational analyses with a close reading of 24 papers.

\subsection{Results}

\paragraph{Where and How Often is Democracy Invoked?}%

\begin{figure}[t!]
    \centering
    \includegraphics[width=\linewidth]{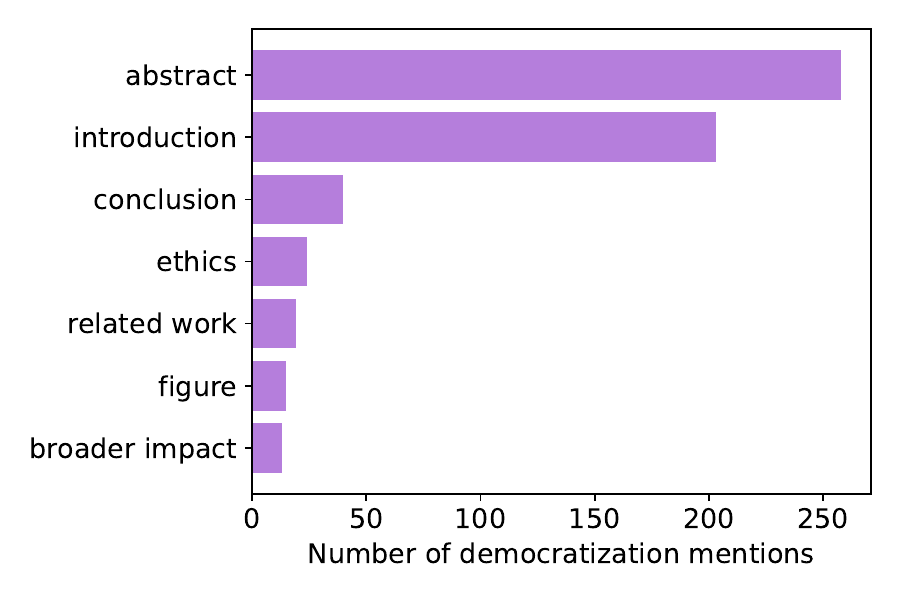}
    \caption{Frequency of paper sections in which mentions of democracy occur.}
    \label{fig:sections}
\end{figure}

We find that the vast majority of papers that mention the ``democra*'' tokens
only mention it once (see \Cref{fig:mentions-per-paper}), and most mentions occur in the abstract, introduction, and conclusion sections (see \Cref{fig:sections}).
These results support our findings in \Cref{sec:democratization-results} that democracy is under-discussed in NLP and ML literature.
Additionally, we find via a close reading of the nine papers with seven or more mentions that a larger number of mentions does \textit{not} necessarily signal higher engagement; for example, mathematical papers frequently refer to ``democratic'' mathematical objects without connecting them to democratic theories.

\paragraph{What Type of Papers are Cited and Why?}

\begin{figure}[t!]
    \centering
    \includegraphics[width=\linewidth]{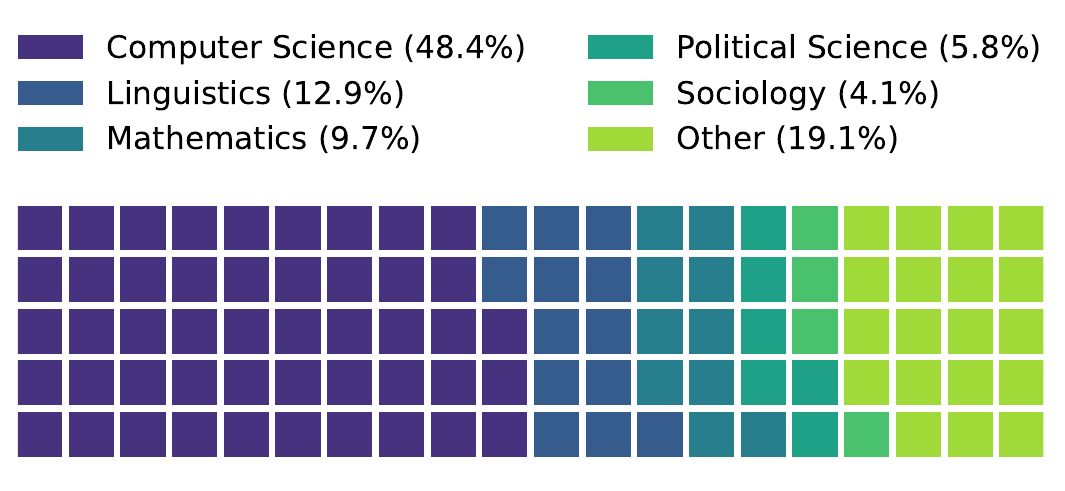}
    \caption{Proportion of fields of study of references cited by papers that mention democracy.}
    \label{fig:fos}
\end{figure}

Our citation analysis reveals that the vast majority of citations are from computer science (48.4\%), which is cited three times more than the second most frequent field, linguistics (12.9\%) (see \Cref{fig:fos}).
Following these, mathematics (9.7\%), political science (5.8\%), and sociology (4.1\%) are most  prominently cited.
Given the direct relationship between ML, computer science and mathematics, and NLP and linguistics, less than 29\% of references are directed towards other disciplines.
This number may not indicate low engagement in and of itself, as it represents an aggregated view of our entire corpus of papers.
However, when we look at individual papers, we find that the majority of them cite zero or one extra-disciplinary works: 181 papers cite \textbf{zero} extra-disciplinary papers, and another 88 cite exactly one.
Given the invocation of far-reaching social concepts such as democracy and democratization, such poor levels of engagement are particularly surprising.
The remaining 220 papers constitute a long-tail that engages more extensively with literature outside of NLP and ML.

Analyzing the extra-disciplinary citations, we see that most citations are from the social sciences, %
especially political science.
However, when considering citational intent, %
we find that most %
(82.3\%) are cited as background, and 15.5\% and 2.14\% are in the context of methods and results, respectively.
Thus, even when work on democracy and democratization consults extra-disciplinary research, it may be primarily used to frame work rather than engage with methods or analyses of results.
Simply citing scholarship on democracy (e.g., ``hit and run'' citations;~\citet{citation-count}) is not equivalent to meaningfully engaging with it.
To evaluate our use of extra-disciplinary citational intent as a proxy for meaningful engagement with democratic theories, we closely read the 15 papers with background, methods, \textit{and} results citations of extra-disciplinary work.
We observe that only the papers citing political science and economics literature, in particular, for methods and results, exhibit deeper consideration of theories of democracy and democratization.
Among citations of political science and economics references, most are still background (84.4\%), with 13.5\% and 2.15\% in the context of methods and results. We confirm that the nine papers with background, methods, \textit{and} results citations of political science and economics literature indeed meaningfully engage with democracy and democratization.

\subsection{Qualitative Examples}
\label{sec:qual-ex}

Below, we present some examples of papers with higher engagement with democratic theories:

\paragraph{Public Dialogue: Analysis of Tolerance in Online Discussions}
\citet{mukherjee-etal-2013-public} discusses tolerance in online discussion, citing work on public spheres, deliberative democracy, tolerance, public dialogue, deliberation, disagreement, and consensus.
The authors ground their methodology in this cited work and perform a ``computational study of tolerance'' in online discussions.
They also interpret their results in the context of this literature, discussing consequences for deliberative discussion and society at large when sustained disagreement turns into intolerance.

\paragraph{A Mathematical Model For Optimal Decisions In A Representative Democracy}
\citet{magdonismail2018} propose a mathematical model for decision-making under representative democracy.
Their extra-disciplinary citations include social sciences and mathematical social sciences.
Through discussion of the differences between direct democracy and representative democracy, the authors motivate a new mathematical model to study the quality-quantity tradeoff with different numbers of representatives, for different types of voting issues, and with different levels of public competence.

\paragraph{Asking Too Much? The Rhetorical Role of Questions in Political Discourse}
\citet{zhang-etal-2017-asking} identify a rich source of rhetorical information in questions in UK parliamentary debates, and thus analyze these as an example of the rhetorical aspects of question in political discourse.
They cite extra-disciplinary literature that establishes the role of questions in democratic processes, and motivate their work as a quantitative examination of aspects that have mostly been qualitatively examined before.
Their unsupervised approach discovers clusters of question types asked in parliamentary discussions.
The authors present an analysis of these clusters and the members of parliament posing these questions (in terms of their tenure and their affiliation to the governing party or the opposition), grounded within extra-disciplinary literature about UK politics and history.

\paragraph{Legal and Political Stance Detection of SCOTUS Language}
\citet{bergam-etal-2022-legal} studies the Supreme Court of the United States (SCOTUS) using text analysis and stance detection on publicly available documents.
Grounding their motivation and analysis in literature about public opinion and democratic principles, the authors also compare their approach to existing metrics from the social sciences, and show how a result about case salience parallels existing findings in political science research.
Finally, they note a trade-off common to the quantitative social sciences in their ethics statement, i.e., that quantitatively analyzing text at scale erases many aspects of its complexity, even as it helps to uncover patterns that cannot feasibly be uncovered by a single qualitative researcher.

\section{Democratic Theories and NLP and ML}
\label{sec:background}

As democratization has had a long history of study %
starting from 1100 BCE in ancient 
Phoenicia~\cite{jacobsen1943primitive,Glassman2017-wv,graeber2021dawn}, in this section, we consider select theories of democracy as a basis for how NLP and ML research has understood and operationalized democracy. %
We argue that these theories %
can provide foundations for more democratic NLP and ML technologies by 
making democratic discussions representative and efficient, diversifying forums for democratic dialogues, and dismantling barriers to participation in democratic processes.

\paragraph{Deliberative Democracies}
Deliberation and inclusion in the democratic process are often highlighted as goals for democratic societies \citep{Roberts2004PublicDeliberation} and technologies \citep{Gilman2023Democratizing}.
Indeed, in our surveyed papers, democratic deliberation often appears as a goal (see \Cref{sec:conceptualizations}).
Deliberative democracy is a form of democracy that emphasizes processes where participants can debate a particular object (e.g., a policy or technology) on its merits and make collective decisions about its implementation~\cite{Goodin_Democratic_2000}.
Deliberative democratic theory %
thus provides an avenue for obtaining more legitimacy of decisions by engaging wider publics in conversation about the use and application of research artifacts~\cite{Rosenberg_Rethinking_2007}.

\paragraph{Democratic Spheres}

As diversity and equal representation are values often associated with democracy in NLP and ML, this raises the question of how we might achieve such goals.
While deliberative democracy provides an avenue for engaging publics, creating a single democratic arena---or sphere, as argued for by \citet{habermas1991structural}---for a large and diverse group gives weight to the loudest voices and majoritarian perspectives.
This risks relegating many communities to the margins, particularly when the publics are large.
In contrast, a plurality of public spheres, which each represent smaller communities, can afford better representation of all communities~\cite{Fraser}.
In practice, if NLP and ML research is consulting a larger group, it can be useful to divide the group into smaller segments, for all voices to be heard.

\paragraph{Democracy and Power}\citet{Mumford_Authoritarian_1964} has argued that 
technology can either afford access, agency, and distribute power, i.e., be democratic, or consolidate power within a small set of actors, i.e., be authoritarian.
Therefore, efforts towards operationalizing the democratization of NLP and ML need to understand and address barriers to public participation and uneven distributions of power.
In relation to discriminatory ML, \citet{Kalluri2020} and \citet{d2020data} have argued that searching for fair ML can serve as a distraction to considering how ML distributes power. %

\paragraph{Radical Egalitarian Democracies}

One approach towards dismantling power differentials and barriers to participation is egalitarian democratic theory.
Under this framework, all humans must have equal access to participate in democratic processes, and these processes should in turn institute programs that dismantle systems of oppression~\cite{Wright2010EnvisioningRU}.
However, the development, operation, and control of NLP and ML technologies are currently determined by the interests of privately held companies~\cite{Zaremba_Dhar_Ahmad_Eloundou_Santurkar_Agarwal_Leung_2023,AnthropicAI,talat-etal-2022-reap,Gray_Widder_West_Whittaker_2023}, under processes that consolidate impact within a small segment of society.
Addressing barriers to public participation and power differentials, as seen through egalitarian democratic theory, would require rethinking processes of public engagement %
in all stages of the development lifecycle.\looseness=-1

\section{Discussion, Conclusion, and Recommendations}

Our thematic and large-scale mixed-methods analyses show that democracy is used in NLP and ML with infrequent operationalization of democratization, vastly different views of what democracy means, and low levels of interdisciplinary engagement.
Overall, our results show that when invoking democracy, NLP and ML researchers only shallowly engage with the centuries of literature from philosophy and social science devoted to it.
It is thus necessary that
NLP and ML researchers describe what they mean by and how they intend to operationalize democratization, to avoid misrepresenting public control of AI and bolstering ``utopian-idealistic'' AI hype~\cite{Sudmann2019}.

In particular, researchers should reflect on what values and concepts they associate with democratization, how their understanding of democratization may be contested, and how their usage of ``democratization'' may be overloaded or overhyped. 
We also echo \citeposs{Seger2023DemocratisingAM} call to simply use the word ``access'' rather than ``normatively loaded language'' like ``democratization'' when discussing access-related questions.
Researchers should go beyond ```access' as the sole condition for participation''~\cite{Luchs2023} and discuss the processes for ``democratic oversight and control'' of their artifacts~\cite{Verdegem2022}.
To this end, they should explicate the causes, targets, methods, and goals of democratization, what it means for their research to be fully democratic, and which opportunities and limits to public participation and control emerge.

Moreover, when invoking democracy and related concepts, researchers should detail how their understanding is informed by underlying theory and ensure to draw from and cite relevant literature.
Conversely, if it is not, they should explicitly indicate this in their work.
In both cases, researchers should reflect on where their conceptualizations fail with respect to their research and goals, and which challenges remain unresolved by their work.
For example, when invoking democratization, researchers should explicitly note what remains unresolved in their goal of democratized technologies.
Some efforts, e.g., OpenAI's call for democratic inputs to AI~\citep{Zaremba_Dhar_Ahmad_Eloundou_Santurkar_Agarwal_Leung_2023} and Anthropic AI's \citet{CIP2024Roadmap}, 
appear to engage more deeply with definitions and implications of democratic AI,
yet do not critically examine questions of power and control.
Similarly, \citet{Djeffal2019} operationalizes AI democratization in line with democratic traditions, including ``parliamentary processes to debate and regulate artificial intelligence.'' 
However, on the whole, we must urgently ``reflect on [our] engagement with other fields'' \citep{wahle-etal-2023-cite}.
While engagement with democratic theory is a necessary precondition for research towards democratizing NLP and ML technologies, it is also necessary to address the hegemonic praxis of NLP and ML, and how it begets or hinders  democratic technologies.

\section*{Limitations}

In our analysis, we may miss relevant NLP and ML literature that treats democratization or democracy due to our focus on the ACL Anthology, ICLR, ICML and NeurIPS.
In our choice of these venues, we are not explicitly controlling for differences in prestige (e.g., workshop papers in the ACL anthology, c.f. main conference papers) or focus (most notably, NLP versus ML), an analysis of which we leave to future work. We further cannot account for the perspectives of NLP and ML researchers who have richer conceptualizations of democratization but are not writing about it.
In addition, our filtering of excerpts based on keywords like ``democra'' may cause us to exclude important discussions of democracy-adjacent concepts that do not use the word.
This may be worsened by parsing errors stemming from our methods and the Semantic Scholar API. The Semantic Scholar API can also fail to correctly predict scholarly metadata, including fields of study and intent, which may affect our results.
Furthermore, our discussion of theories of democracy (see \Cref{sec:background}) is far from exhaustive, given the rich history of the subject.

\section*{Ethical Considerations}

Our paper emphasizes careful consideration and usage of the term ``democratization,'' especially given its relation to democracy, and urges drawing from extra-disciplinary literature on democratic theories.
This is important for accurately representing the distribution of power, public control, and progress in NLP and ML.
In light of our findings, we stress that our analysis only captures a snapshot in time and that researchers' perspectives on democratization and democracy can evolve; moreover, the text of papers may not wholly reflect the perspectives of their authors, given the diversity of opinions among authors and reviewing incentives.

\section*{Acknowledgements}
We thank the anonymous reviewers for their insightful feedback. We also greatly appreciate Luca Soldaini, Lucy Li, Maria Antoniak, and Shaily Bhatt for their constructive comments on the presentation and organization of the paper. We further thank Shreya Chowdhary and Skyler Wang for preliminary discussions about this project, and Luca Soldaini for help with Semantic Scholar.

\bibliography{anthology,custom}
\bibliographystyle{acl_natbib}

\clearpage

\appendix
\onecolumn

\section{Methodological Details}
\label{sec:methodological-details}

\begin{figure}[t!]
    \centering
    \includegraphics[width=\linewidth]{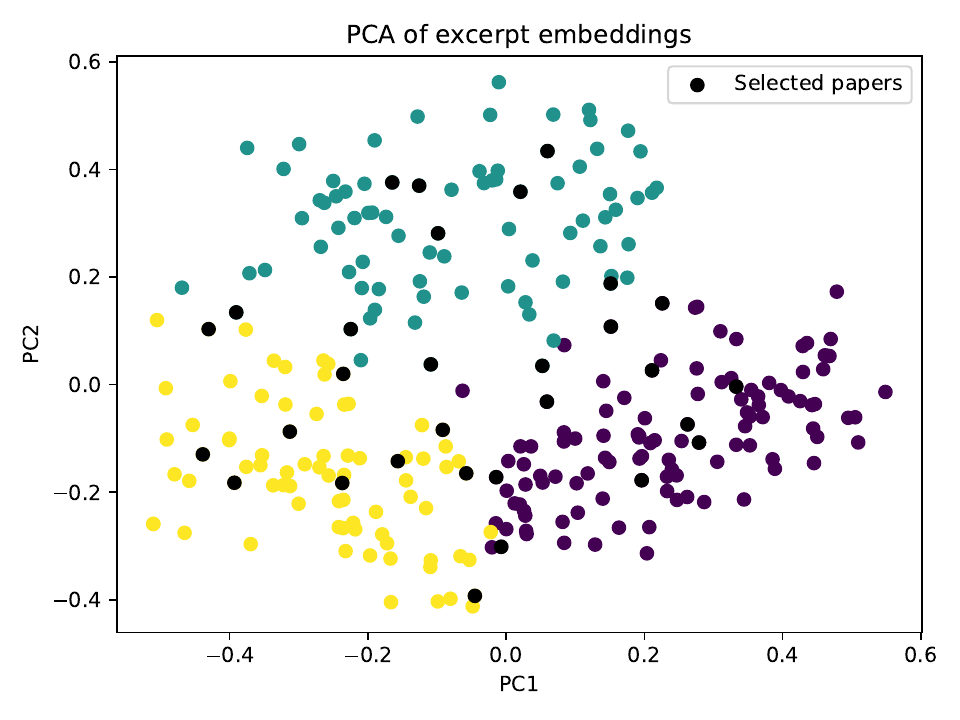}
    \caption{PCA and spectral clustering of excerpt embeddings, along with selected papers. Points that are the same color belong to the same cluster.}
    \label{fig:clustering}
\end{figure}

\Cref{tab:democra-false-pos} lists all false positive terms that we use in our first stage of manual filtering. \Cref{fig:clustering} shows the results of our PCA and clustering of embedded excerpts, with the darkest colour indicating the papers we select for reading and annotating fully.

\begin{table*}[ht!]
    \centering
    \resizebox{\textwidth}{!}{  
        \begin{tabular}{lll}
        \toprule
        democrat &  Democratic National Committee &                  Project ANR Democrat                                    \\
                                  democrats &              Liberal Democratic Party &                                                             Democrat system \\
                       Republican Democrat &                      Democratic Party & Description, Modélisation et Détection Automatique Des Chaînes de Référence \\
                       Democrat Republican &            German Democratic Republic &                                                                    DEMOCRAT \\
                   Republican and Democrat &            Getman Democratic Republic &                                                                  Democratic \\
                   Democrat and Republican & Democratic People's Republic of Korea &                                    christian democratic parliamentary group \\
                 Republicans and Democrats &            Christian Democratic Union &                                                                  \#democracy \\
                 Democrats and Republicans &                   Democratic Alliance &                                                             Democracy party \\
                    Republican or Democrat &               United Democratic Front &                                  Democrazia Cristiana / Christian Democracy \\
                    Democrat or Republican &      Democratic Governors Association &                                                           \#democratic\_party \\
                  Republicans or Democrats &                 China Democracy Party &                                           social-democratic political party \\
                  Democrats or Republicans &                    Christian Democrat &                                                    social-democratic leader \\
           the Republican and the Democrat &                    Democratic primary &                                              Center for Media and Democracy \\
           the Democrat and the Republican &                  Democratic primaries &                                              democratic president candidate \\
         the Republicans and the Democrats &               Somali Democratic Party &               Stichting Democratie and Media (Democracy \& Media Foundation) \\
         the Democrats and the Republicans &                  New Democratic Party &                                        Swedish social democratic politician \\
            the Republican or the Democrat &            Democratic Socialist Party &                                                      democratic congressman \\
            the Democrat or the Republican &                      Liberal Democrat &                                                  social democratic movement \\
          the Republicans or the Democrats &              Democratic Left Alliance &                                                        Christian democratic \\
          the Democrats or the Republicans &        Alliance for Democracy in Mali &                              social democratic, centre-left political party \\
         democratic and republican parties &              Syrian Democratic Forces &                                                     Democratic Labour Party \\
                 Democratic Party of Japan &                        Democracy Now! &                                              democratic republic of germany \\
         Liberal Democratic Party of Japan &        Movement for Democratic Change &                      Historical Press of the German Social Democracy Online \\
                   Social Democratic Party &                        Democracy Week &                                Forum voor Democratie, 'Forum for Democracy' \\
                      Democratic candidate &                 Democratic-controlled &                                            centre-right party New Democracy \\
                     Democratic candidates &             Croatian Democratic Union &                                                         Partito Democratico \\
          Democratic republic of the Congo &                 Kurd Democratic Party &                                                        Social Democracy (S) \\
         Democratic presidential candidate &                  New Democratic Union &                                       Forum Migration and Democracy (MIDEM) \\
        Democratic presidential candidates &                          ANR Democrat &                                                                             \\
        \bottomrule
        \end{tabular}
    }
    \caption{False positives when matching ``democra*'' in corpus.}
    \label{tab:democra-false-pos}
\end{table*}

\section{Additional Results}
\label{sec:additional-results}

\subsection{All concepts and values}

\begin{figure*}[t!]
    \centering
    \includegraphics[width=0.45\linewidth]{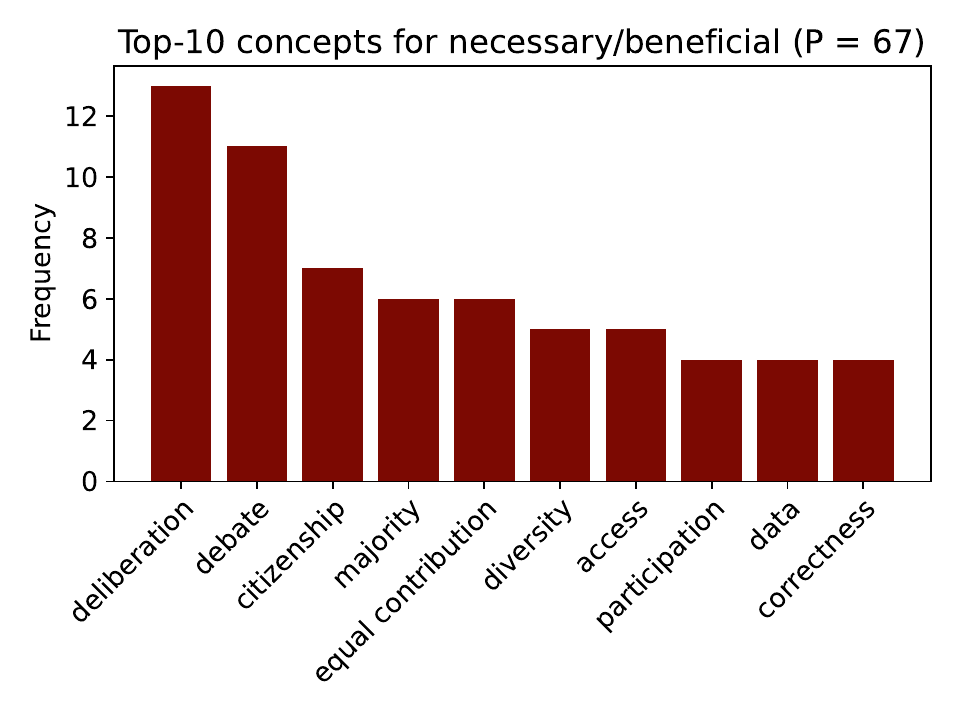}
    \includegraphics[width=0.45\linewidth]{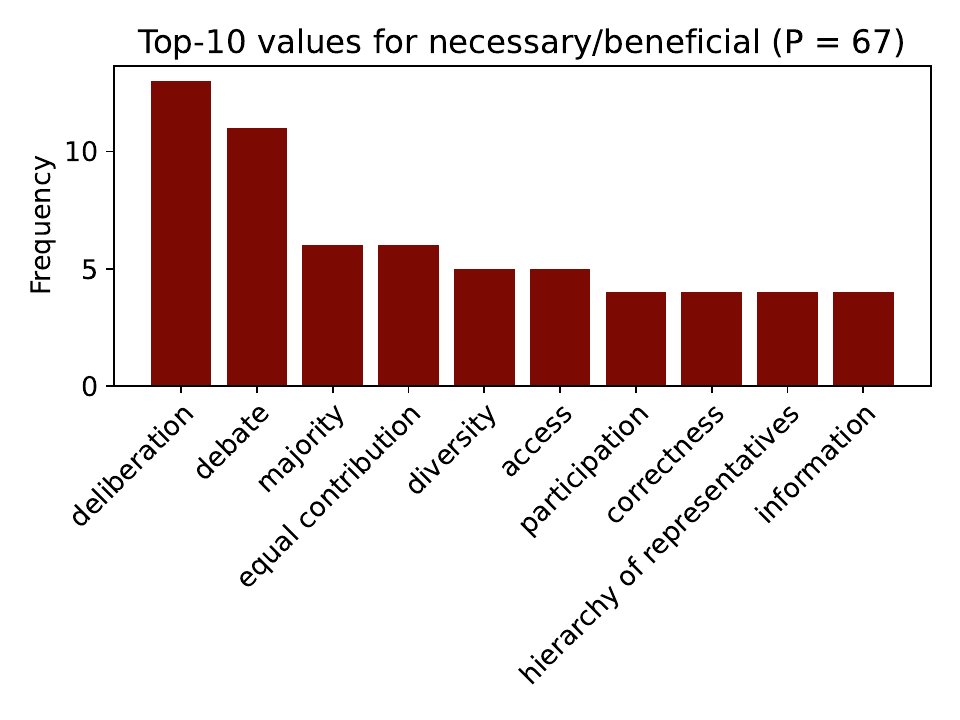}
    \includegraphics[width=0.45\linewidth]{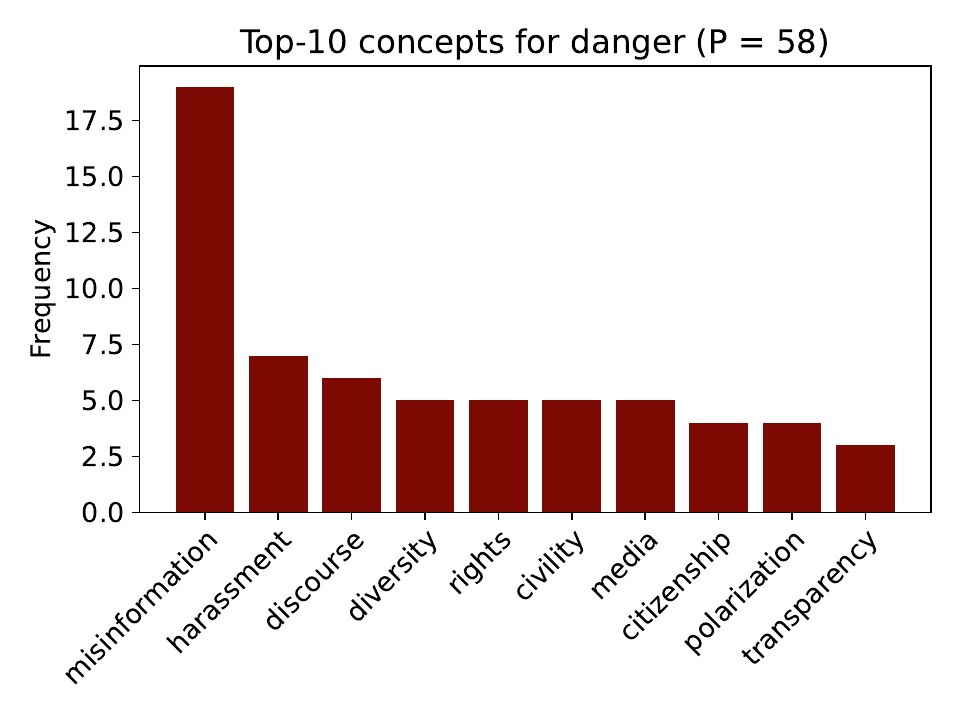}
    \includegraphics[width=0.45\linewidth]{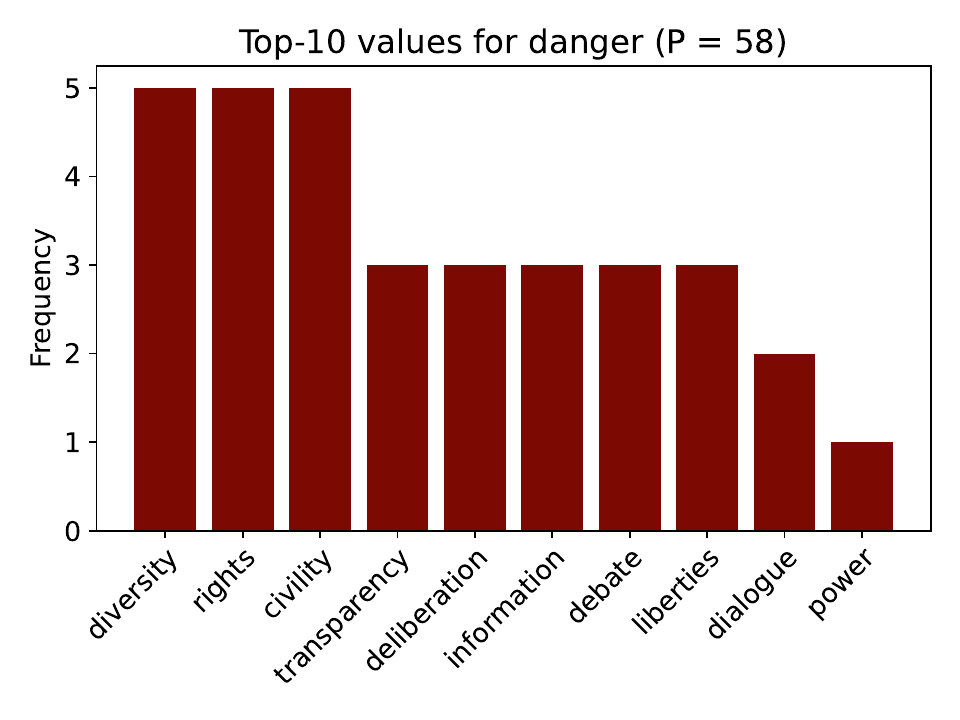}
    \includegraphics[width=0.45\linewidth]{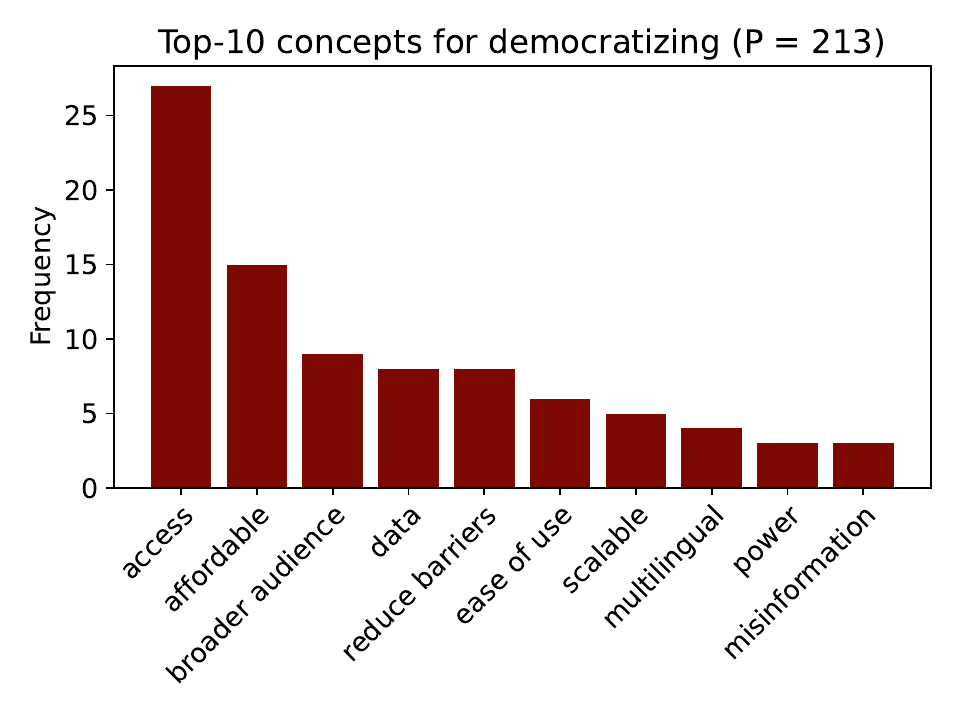}
    \includegraphics[width=0.45\linewidth]{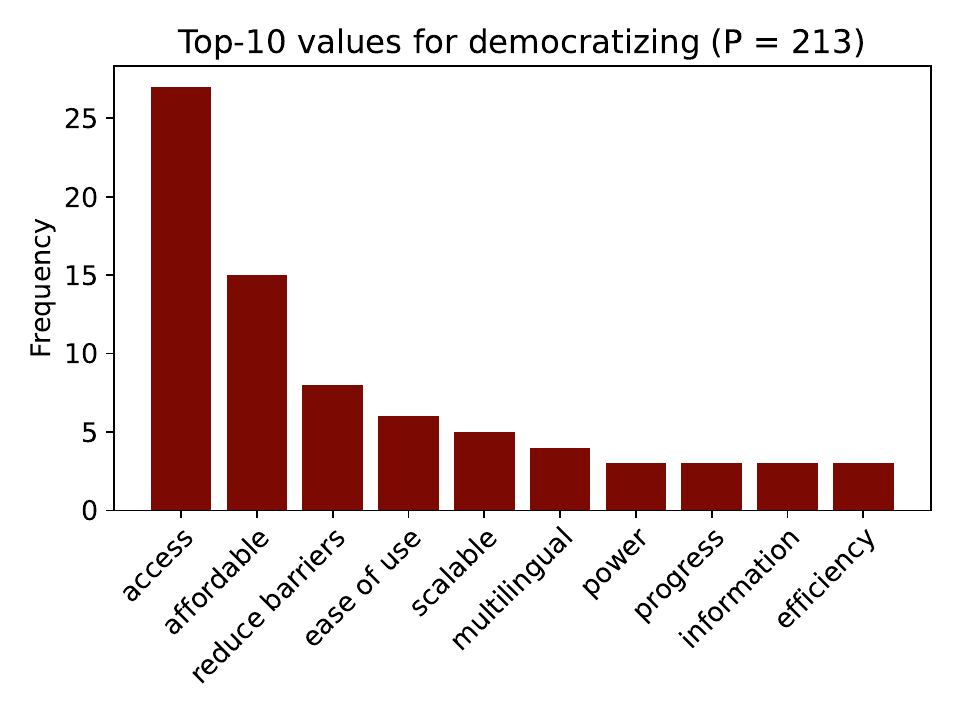}
    \includegraphics[width=0.45\linewidth]{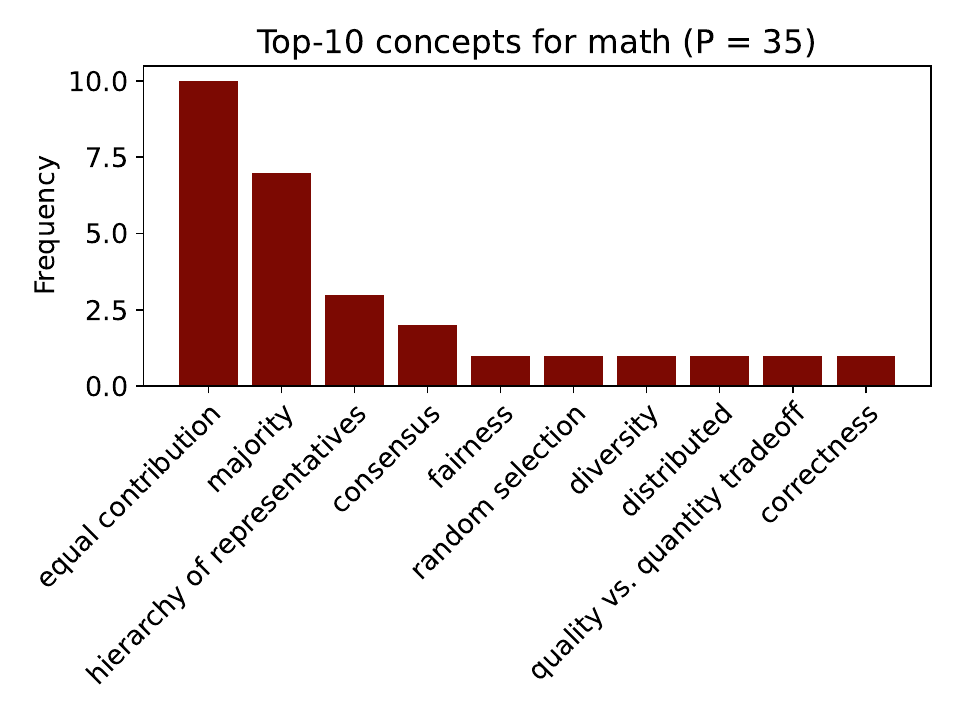}
    \includegraphics[width=0.45\linewidth]{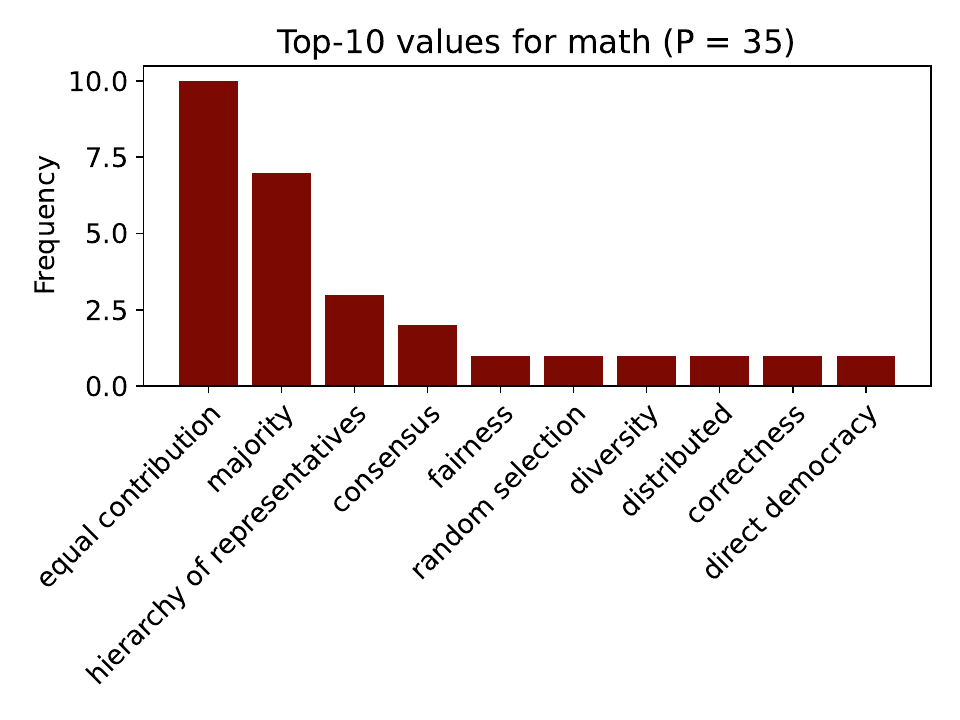}
    \caption{Frequency of concepts (left) and values (right) associated with democracy in papers, stratified by paper themes. For each theme, $P$ refers to the number of papers annotated as having that type of theme.}
    \label{fig:concepts-values-stratified}
\end{figure*}

Values are understood to be a subset to concepts, which are regarded explicitly or implicitly as pertinent to a specific context (e.g., in relation to democracy).
The authors undertake a subjective, direct democratic process to distinguish concepts from values.
Tables \ref{tab:all-concepts} and \ref{tab:all-values} shows all concepts and values we find during excerpt annotation. \Cref{fig:concepts-values-stratified} shows the top concepts and values for each theme.

\subsection{Where do (extra-disciplinary) references come from?}

When considering where references are published, we find that the top five venues of all references are: NLP venues (*CL conferences), machine learning venues (ICLR/ICML/NeurIPS), arXiv, computer vision venues (CVPR/ICCV/ECCV), and the AAAI Conference on Artificial Intelligence.
In contrast to this general pattern, extra-disciplinary references are mostly published in political science and social science journals, i.e., the American Political Science Review, Political Analysis, Nature, PloS ONE, Social Science Research Network, Science, and so on.
The most frequently cited extra-disciplinary references are typically cited for methods, e.g., content analysis, agreement computations, discourse network analysis, or related to fake news and polarization. The most cited extra-disciplinary references in our corpus are:
\begin{enumerate}
    \item Text as Data: The Promise and Pitfalls of Automatic Content Analysis Methods for Political Texts
    \item A Coefficient of Agreement for Nominal Scales
    \item Social Media and Fake News in the 2016 Election
    \item Extracting Policy Positions from Political Texts Using Words as Data
    \item Discourse Network Analysis: Policy Debates as Dynamic Networks
    \item Measuring Political Deliberation: A Discourse Quality Index
    \item Discourse Coalitions and the Institutionalization of Practice: The Case of Acid Rain in Great Britain
    \item CUNY Academic Works
    \item Exposure to Opposing Views on Social Media Can Increase Political Polarization
\end{enumerate}

\begin{table*}[ht!]
    \centering
    \resizebox{0.5\columnwidth}{!}{  
        \begin{tabular}{lll}
        \toprule
        generalizability & protection & dialogue \\
        literacy & debate & decentralization \\
        public opinion & freedom & sustainability \\
        fairness & moderation & emotion \\
        WEIRD & replicability & justice \\
        liberties & environment & voting \\
        anti-power & integrity & citizenship \\
        equal contribution & resource-efficient & low-resource \\
        interaction & engagement & broader audience \\
        hierarchy of representatives & multilingual & scalable \\
        rights & news & efficiency \\
        governance & transparency & caution \\
        acceleration & disagreement & civility \\
        reduce barriers & protest & anxiety \\
        discrimination & progress & data \\
        translation & quality & access \\
        happiness & reasoning & power \\
        constitution & harassment & accountability \\
        questioning & majority & consistency \\
        competence & value & social good \\
        reflection & open-source & cohesion \\
        equal representation & evolving & polarization \\
        informed & argument & campaign \\
        fast & available & cooperation \\
        representation & trust & information \\
        responsibility & random selection & inclusion \\
        diversity & quality vs. quantity tradeoff & direct democracy \\
        political party & election & bill writing \\
        correctness & affordable & choice \\
        conflict & ease of use & discourse \\
        equality & distributed & media \\
        education & misinformation & discussion \\
        privacy & participation & propaganda \\
        complexity & critical & benefit \\
        proficiency & censorship & AI \\
        rational & consensus & lack of prejudice \\
        disinformation & deliberation &  \\
        \bottomrule
        \end{tabular}
    }
    \caption{All associated concepts found when annotating excerpts.}
    \label{tab:all-concepts}
\end{table*}

\begin{table*}[ht!]
    \centering
    \resizebox{0.5\columnwidth}{!}{  
        \begin{tabular}{lll}
        \toprule
          sustainability &                 disagreement &         moderation \\
                fairness &                      caution &    reduce barriers \\
                argument &                       choice &            justice \\
                progress &                   optimality &   direct democracy \\
                   trust &                participation &           rational \\
        random selection &                  proficiency & resource-efficient \\
               consensus &                    inclusion &          diversity \\
               available &                     critical &          liberties \\
            multilingual &                   engagement &        cooperation \\
               reasoning &                  interaction &         efficiency \\
        generalizability &                      benefit &        open-source \\
               integrity &               accountability &         reflection \\
                literacy &                 transparency &             access \\
             social good &                     evolving &   decentralization \\
                civility &                     cohesion &           informed \\
                conflict &         equal representation & equal contribution \\
                majority &                replicability &     representation \\
             correctness &                     equality &             debate \\
                 privacy &                        power &        distributed \\
                 quality & hierarchy of representatives &         protection \\
            deliberation &            lack of prejudice &         affordable \\
             information &                       rights &         discussion \\
             ease of use &                     dialogue &          happiness \\
          responsibility &                         fast &         anti-power \\
               education &                        value &        consistency \\
                scalable &                   competence &                    \\
        \bottomrule
        \end{tabular}
    }
    \caption{All associated values found when annotating excerpts.}
    \label{tab:all-values}
\end{table*}

\end{document}